\documentclass[runningheads]{llncs}
\usepackage{graphicx}

\usepackage{tikz}
\usepackage{comment}
\usepackage{amsmath,amssymb} %
\usepackage{color}
\usepackage{graphicx}
\usepackage{amsmath}
\usepackage{amssymb}
\usepackage{booktabs}

\usepackage{epsfig}
\usepackage{graphicx}
 
\usepackage{graphics} %

\usepackage{amssymb}  %
\usepackage{makecell}
\usepackage{cite}
\usepackage{booktabs}
\usepackage{amsfonts,amssymb}
\usepackage{textcase}
\usepackage[tablename=TABLE]{caption}
\usepackage{booktabs}
\usepackage{placeins}
\usepackage{multicol}
\usepackage{multirow}
\usepackage{textcomp}
\usepackage{textcase}
\usepackage[tablename=TABLE]{caption}

\usepackage{enumitem}
\usepackage{diagbox}
\usepackage{caption}
\usepackage{float}
\usepackage[toc]{appendix} 
\usepackage{gensymb}
\usepackage{mathtools}
\usepackage{caption}
\usepackage{subcaption}
\usepackage{capt-of}
\usepackage{array}
    \newcolumntype{P}[1]{>{\centering\arraybackslash}p{#1}}
\usepackage{graphicx}
\usepackage{caption}
\DeclareCaptionLabelFormat{andtable}{#1~#2  \&  \tablename~\thetable}
\usepackage[export]{adjustbox}
\usepackage[accsupp]{axessibility}  %
\usepackage[pagebackref,breaklinks,colorlinks]{hyperref}
\usepackage{array}
\newcommand{\PreserveBackslash}[1]{\let\temp=\\#1\let\\=\temp}
\newcolumntype{C}[1]{>{\PreserveBackslash\centering}p{#1}}
\newcolumntype{R}[1]{>{\PreserveBackslash\raggedleft}p{#1}}
\newcolumntype{L}[1]{>{\PreserveBackslash\raggedright}p{#1}}
\usepackage{xspace}
\makeatletter
\DeclareRobustCommand\onedot{\futurelet\@let@token\@onedot}
\def\@onedot{\ifx\@let@token.\else.\null\fi\xspace}

\def\eg{\emph{e.g}\onedot} 
\def\ie{\emph{i.e}\onedot}

\def\wrt{w.r.t\onedot} 
\def\etal{\emph{et al}\onedot}
\makeatother
\usepackage{booktabs}
\usepackage{caption}
\makeatletter
\newcommand{\currentfsize}{\f@size pt}
\makeatother

\newdimen\fsize
\newcommand{\setfsize}{\setlength{\fsize}{\currentfsize}}
\usepackage[capitalize]{cleveref}
\crefname{section}{Sec.}{Secs.}
\Crefname{section}{Section}{Sections}
\Crefname{table}{Table}{Tables}
\crefname{table}{Tab.}{Tabs.}

\newcommand{\ba}{\mathbf{a}}

\newcommand{\bI}{\mathbf{I}}

\newcommand{\bo}{\mathbf{o}}

\newcommand{\bq}{\mathbf{q}}
\newcommand{\bR}{\mathbf{R}}

\newcommand{\bt}{\mathbf{t}}

\newcommand{\bz}{\mathbf{z}}

\newcommand{\btheta}{\boldsymbol{\theta}}

\newcommand{\nR}{\mathbb{R}}

\newcommand{\cL}{\mathcal{L}}

\newcommand{\figref}[1]{Fig.~\ref{#1}}
\newcommand{\secref}[1]{Section~\ref{#1}}

\newcommand{\tabref}[1]{Table~\ref{#1}}

\makeatletter
\DeclareRobustCommand\onedot{\futurelet\@let@token\@onedot}
\def\@onedot{\ifx\@let@token.\else.\null\fi\xspace}
\def\eg{e.g\onedot} 
\def\ie{i.e\onedot}

\def\wrt{wrt\onedot}

\def\etal{et~al\onedot}

\makeatother

\newcommand{\boldparagraph}[1]{\vspace{0.2cm}\noindent{\bf #1:} }

\makeatletter
\newcommand{\printfnsymbol}[1]{%
  \textsuperscript{\@fnsymbol{#1}}%
}
\makeatother

\begin{document}
\pagestyle{headings}
\mainmatter
\def\ECCVSubNumber{4677}  %

\title{A Visual Navigation Perspective for Category-Level Object Pose Estimation} %

\makeatletter
\def\blfootnote{\gdef\@thefnmark{}\@footnotetext}
\makeatother

\titlerunning{A Visual Navigation Perspective for Category-Level Object Pose Estimation}
\author{Jiaxin Guo\inst{1,2} \and
Fangxun Zhong\inst{2} \and
Rong Xiong\inst{1}  \and
Yunhui Liu\inst{2} \and \\
Yue Wang\inst{1}$^{,*}$%
\and
Yiyi Liao\inst{1}$^{,**}$%
}
\authorrunning{J. Guo et al.}
\institute{Zhejiang University, Hangzhou, China \and
The Chinese University of Hong Kong, Hong Kong, China\\
\email{\{jxguo, fxzhong, yhliu\}@mae.cuhk.edu.hk, } \\ \email{\{rxiong, ywang24, yiyi.liao\}@zju.edu.cn}}
\maketitle

\begin{abstract}

This paper studies category-level object pose estimation based on a single monocular image. Recent advances in pose-aware generative models have paved the way for addressing this challenging task using analysis-by-synthesis. The idea is to sequentially update a set of latent variables, \eg, pose, shape, and appearance, of the generative model until the generated image best agrees with the observation. However, convergence and efficiency are two challenges of this inference procedure. In this paper, we take a deeper look at the inference of analysis-by-synthesis from the perspective of visual navigation, and investigate what is a good navigation policy for this specific task. We evaluate three different strategies, including gradient descent, reinforcement learning and imitation learning, via thorough comparisons in terms of convergence, robustness and efficiency. Moreover, we show that a simple hybrid approach leads to an effective and efficient solution. We further compare these strategies to state-of-the-art methods, and demonstrate superior performance on synthetic and real-world datasets leveraging off-the-shelf pose-aware generative models.
\keywords{category-level object pose estimation, analysis-by-synthesis}
\end{abstract}

\blfootnote{$^*$ Co-corresponding author. $^{**}$ Corresponding author.}

\section{Introduction}

\begin{figure}[t]
\centering
\includegraphics[width=0.9\linewidth]{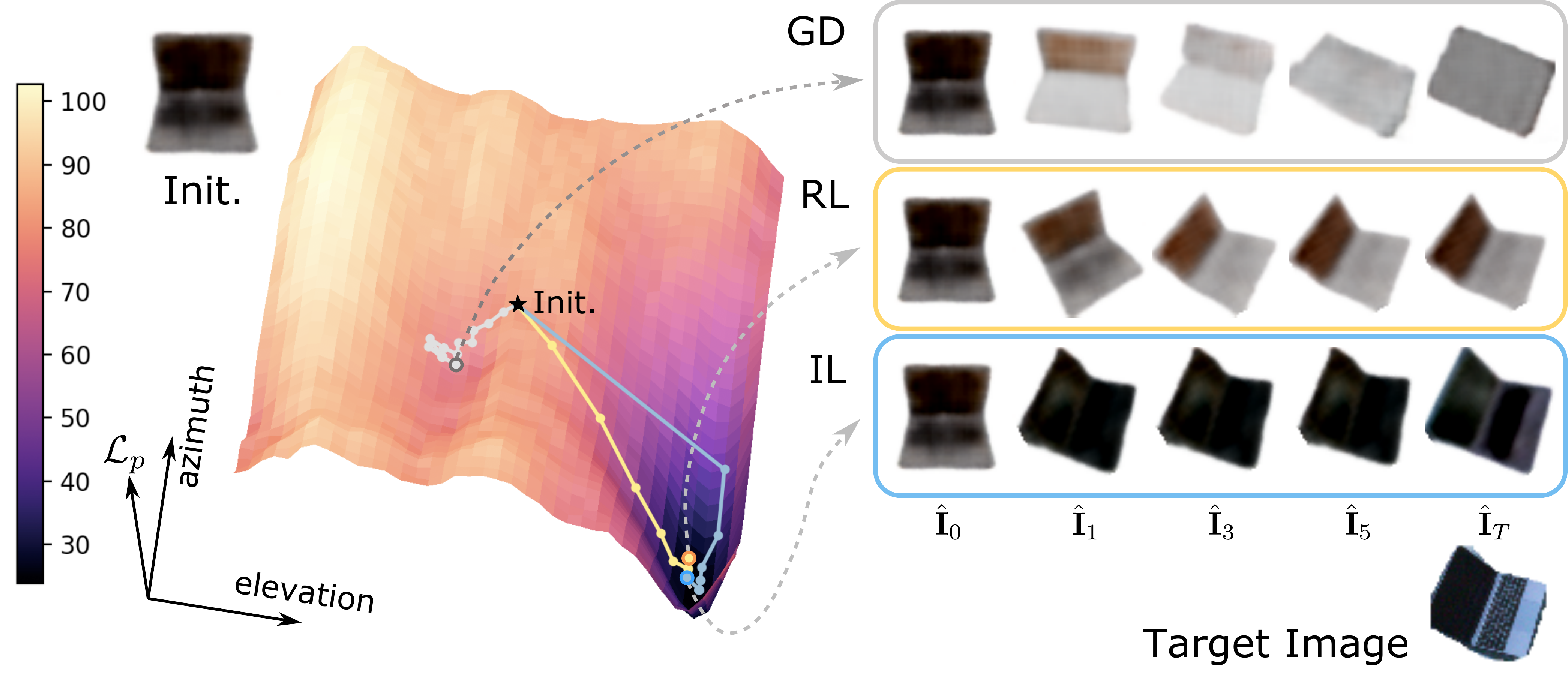}
\caption{\textbf{Inference of Analysis-by-Synthesis.} We illustrate the perceptual loss between the synthesized image and the target image, calculated over a grid of azimuth and elevation. We further show the navigation trajectory of gradient descent (GD), reinforcement learning (RL) and imitation learning (IL) given the same initialization, including the synthesized images generated at multiple steps. Note that GD converges to a local minimum due to the non-convex loss landscape. RL and IL converge in the correct direction while IL converges faster. }
\label{fig:teaser}
\end{figure}

Object pose estimation is a fundamental research problem that aims to estimate the 6 DoF pose of an object from a given observation. To enable broad applications in augmented reality and robotics, it is essential that the object pose estimation methods allow for generalizing to unseen objects and being applicable to widely used sensors, \eg, monocular cameras. Thus, there is a growing interest in the challenging task of category-level object pose estimation based on a single monocular image \cite{chen2020category}.

As a classic idea in computer vision, analysis-by-synthesis has recently shown competitive performance in object pose estimation~\cite{Krull2015ICCV,Shao2020CVPR,chen2020category,yen2020inerf}. This line of approaches leverages a forward synthesis model that can be controlled by a low-dimensional input, \eg, object pose, and infers the pose via render-and-compare. Given an observation image,  multiple images can be synthesized under different object poses, and the one that best matches the observation is selected. While earlier methods only apply to instances of known CAD models taking a graphics renderer as the forward model ~\cite{Krull2015ICCV,Shao2020CVPR}, recent works extend this idea to category-level object pose estimation leveraging pose-aware generative models and demonstrate superior performance compared to direct pose regression~\cite{chen2020category,yen2020inerf}.

In this paper, we advocate analysis-by-synthesis but identify a major limitation of existing approaches: it is non-trivial to efficiently retrieve the pose that best reproduces the target observation. \figref{fig:teaser} illustrates that existing methods based on gradient descent (GD) are sensitive to initialization and are prone to convergence problems. This is due to the fact that the objective function, \ie, the difference between the synthesized image and the observation, is highly non-convex. Leveraging multiple initial poses is a common remedy for this problem~\cite{park2020latentfusion,chen2020category}. However, this is time-consuming and computationally expensive.

Intending to analyze and improve the \textit{inference} process of analysis-by-synthesis, we view the inference as a visual navigation task, where an agent uses visual input to take a sequence of actions to reach its own goal~\cite{Zhu2017ICRA}. This perspective allows us to take inspirations from the visual navigation literature and compare different navigation policies.
This formulation leads to our main question:
\textit{what is a good navigation policy for category-level object pose estimation?}

To answer this question, we systematically compare different navigation policies. 
Taking the pose-aware generative model as a \textit{simulator}, we explore common strategies in visual navigation, including reinforcement learning (RL)~\cite{Zhu2017ICRA, Mirowski2017ICLR,Pfeiffer2018RAL} and imitation learning (IL)~\cite{Ross2013ICRA,Kretzschmar2016IJRR}. We also study the behavior of GD as a one-step greedy strategy within the same framework.
Specifically, we first investigate how design choices, \ie, planning horizon and loss function, affect the behavior of navigation policies. Next, we compare all strategies \wrt convergence, robustness, and efficiency and make the following observations: 1) Both RL and IL remarkably alleviate convergence problems compared to GD as shown in \figref{fig:teaser}. Despite easily getting stuck in local minima, GD yields a more precise prediction given a good initialization;  2) GD tends to be more robust against disturbance of brightness and shift on the target image;  3) Both RL and IL are more efficient than GD during inference. Compared to RL, IL requires less training time but is less competitive when trained with off-policy data. However, IL achieves similar or even better performance than RL when augmented with on-policy data. Based on these observations, we suggest to combine IL's convergence and efficiency with GD's precision and robustness. We demonstrate that this simple hybrid approach achieves superior performance on category-level pose estimation.

We summarize our contributions as follows: i) We propose to view the inference process of analysis-by-synthesis as a visual navigation task, leveraging the pose-aware generative model as a simulator to provide training data without manual labeling.
ii)  We conduct thorough comparisons between GD, RL, and IL in terms of  convergence, robustness and efficiency. Based on our observations we suggest a simple combination of IL and GD that is effective and efficient.
iii)  We compare different strategies to state-of-the-art methods on category-level object pose estimation on synthetic and real-world datasets. Our hybrid approach shows competitive performance and consistently improves the inference process of different pose-aware generative models. Our code is released at \url{https://github.com/wrld/visual_navigation_pose_estimation.git}.

\section{Related Work}
\boldparagraph{Object Pose Estimation} 
Extensive studies have been conducted for object pose estimation of known \textit{instances}~\cite{Xiang2018RSS,Do2018BMVC,Kehl2017ICCV,drost2010model,li2018deepim,munoz2016fast,Li2019ICCV,park2019pix2pose,hu2019segmentation,Peng2019CVPR,Krull2017CVPR}. Only recently, there has been a growing interest in a more general task of \textit{category-level} 6 DoF object pose estimation for unseen instances in a specific category~\cite{Sahin2018ECCVW,wang2019normalized,Tian2020ECCV,Chen2020CVPR,Wang2021IROS,Chen2021CVPR}. These methods achieve promising results via establishing correspondences across different objects~\cite{wang2019normalized,Tian2020ECCV,Wang2021IROS} or direct regression~\cite{Chen2020CVPR}. In this paper, we are interested in category-level object pose estimation leveraging a pose-aware generative model, eliminating the need for intermediate correspondences compared to ~\cite{wang2019normalized,Tian2020ECCV,Wang2021IROS}. In contrast to direct regression methods~\cite{Chen2020CVPR}, we model the problem as a long-horizon navigation task to approach the goal sequentially via a set of relative updates. Moreover, all aforementioned methods for category-level object pose estimation are applied to RGB-D images while we focus on a single RGB image-based solution.

\boldparagraph{Analysis-by-Synthesis for Object Pose Estimation} %
It is a classical idea to analyze a signal by reproducing it using a synthesis model, which is referred to as analysis-by-synthesis~\cite{Vuille2006Vision}. It has been successfully applied to many tasks, including human pose estimation~\cite{Loper2014ECCV}, object recognition~\cite{Hejrati2014CVPR}, and scene understanding~\cite{Isola2013ICCV,Moreno2016ECCVW}. A few methods leverage this idea for instance-level object pose estimation~\cite{Krull2015ICCV,Shao2020CVPR}, but are not applicable to unseen instances. %

With the rapid progress of pose-aware generative models that allow for generating 2D images under controllable object poses~\cite{Nguyen2019ICCV,Henzler2019ICCV,Liao2020CVPR,schwarz2020graf,Chan2021CVPR,Niemeyer2021CVPR}, analysis-by-synthesis approaches are extended to category-level object pose estimation recently \cite{chen2020category, yen2020inerf, park2020latentfusion}. Among them, a few works demonstrate promising results using a single RGB image~\cite{chen2020category, yen2020inerf}. As the generative models are differentiable, all these approaches leverage gradient descent to infer the object pose. Due to the non-convex objective function, gradient-based optimization suffers from convergence problems. While iNeRF~\cite{yen2020inerf} constrains the initialization range during inference, LatentFusion~\cite{park2020latentfusion} and Chen \etal~\cite{chen2020category} start from multiple pose candidates and keep the one that best aligns with the observation. However, it is computationally expensive, and the computing time increases \wrt the number of initial poses. Another idea is to leverage an encoder to provide a better initialization~\cite{chen2020category,duggal2021secrets}. Note that the common underlying idea of these methods is to improve the initialization. In contrast, we focus on analyzing and improving the policy for updating the pose.

\boldparagraph{Visual Navigation}
By sufficiently interacting with the simulation environment, RL has demonstrated superior performance in long-horizon decision tasks in visual navigation~\cite{Zhu2017ICRA, Mirowski2017ICLR,Pfeiffer2018RAL}.
IL is also commonly adopted when expert demonstrations are available~\cite{Ross2013ICRA,Kretzschmar2016IJRR,Bojarski2016ARXIV}.
In contrast to all aforementioned methods, we propose to adopt a pose-aware generative model as a simulator, where the agent navigates in the input space of a pose-aware generative model for category-level pose estimation. Exploiting RL/IL to learn the gradient is similar to meta gradient descent methods~\cite{andrychowicz2016learning}. Compared to existing meta-GD methods, we utilize the simulator to provide explicit supervision towards the global optimum.

\boldparagraph{Inversion of Generative Models} The idea of analysis-by-synthesis is also closely related to inversion of generative models~\cite{Zhu2016ECCV,Abdal2019ICCV,Shen2020CVPR}, see~\cite{Xia2021ARXIV} for a detailed survey. While all these works focus on enabling the editing of a real image by searching its corresponding latent code, we leverage pose-aware generative models to estimate category-level object poses. In our task, the inverting process is more sensitive to initialization and prone to convergence problems.

\section{Object Pose Estimation as Visual Navigation}

Our goal is to improve the inference procedure of the analysis-by-synthesis pipeline for category-level object pose estimation. In the following, we first formulate the problem as a visual navigation task in \secref{sec:formulation}. Next, we present several navigation policies, including gradient descent (\secref{sec:gd}), reinforcement learning (\secref{sec:rl}) and imitation learning (\secref{sec:il}). 

\begin{figure*}[t]
\begin{subfigure}[t]{\linewidth}
\centering
\includegraphics[width=\textwidth]{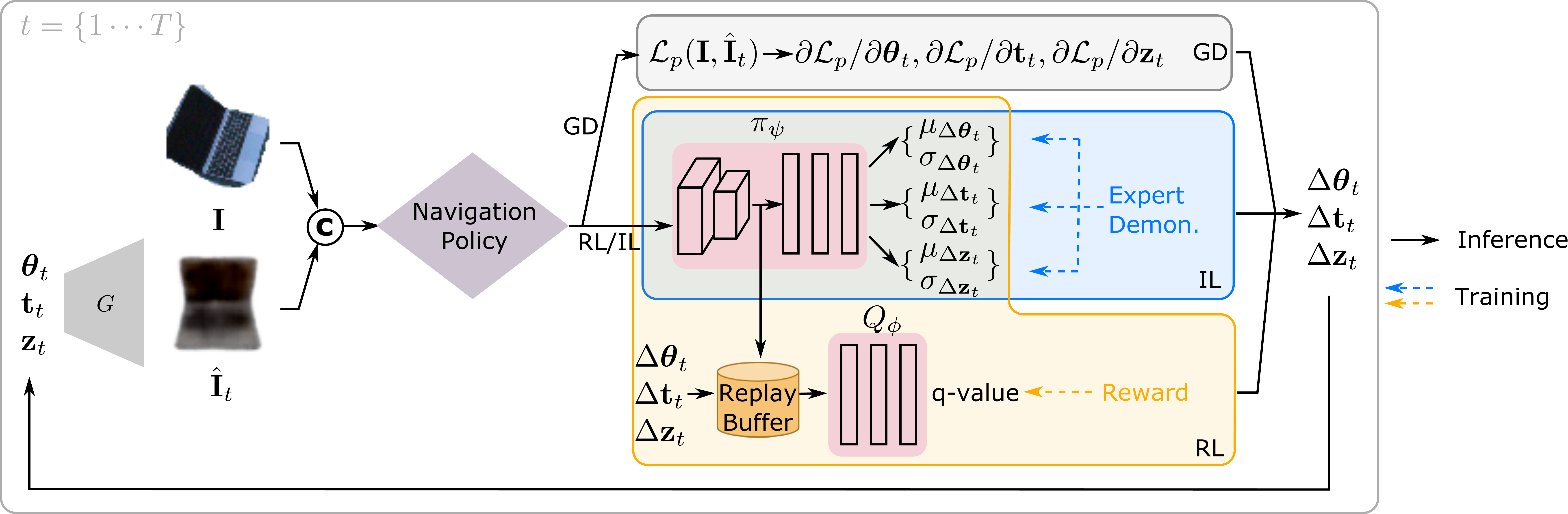}
\end{subfigure}
\caption{\textbf{Category-Level Object Pose Estimation as Visual Navigation.} We illustrate the visual navigation pipeline, where an agent uses visual input (the synthesized image $\hat{\bI}$) to reach a target state $\btheta^*,\bt^*,\bz^*$ via iteratively taking $T$ steps of actions. At a time step $t$, the navigation policy takes as input the synthesized image $\hat{\bI}_t=G(\btheta_t, \bt_t, \bz_t)$ and the target image $\bI$, to update the state $\btheta_t, \bt_t, \bz_t$ via $\Delta\bR_t, \Delta\bt_t, \Delta\bz_t$. We evaluate and compare three different strategies as the navigation policy, including gradient descent (GD), reinforcement learning (RL) and imitation learning (IL). Note that $G$ is fixed and the GD policy does not contain any trainable parameters. Both RL and IL learn the policy via a network parameterized by $\psi$, while supervised by different signals.}
\label{fig:pipeline}
\end{figure*}

\subsection{Problem Formulation}
\label{sec:formulation}
We aim for 6 DoF category-level object pose estimation from a single image via analysis-by-synthesis. The idea is to sequentially update a set of input variables, \eg, object pose, shape and appearance, of a forward synthesis model, until the generated image best matches the target. The corresponding pose is then selected as the prediction.
We view this sequential procedure as a long-horizon visual navigation task as illustrated in \figref{fig:pipeline}. Formally, we model this problem as a Markov decision process (MDP).  Let  $\bI$ denote the target image, and $\hat \bI=G(\btheta, \bt, \bz)$ denote a synthesized image generated by a pose-aware generative model $G$. Here, $G$ takes as input the object's rotation $\bR_{\btheta}\in SO(3)$, which is parametrized using Euler angles $\btheta = [\btheta_x, \btheta_y, \btheta_z]$, translation $\bt\in \nR^3$, and a latent code $\bz$ for its shape and appearance. Note that $G$ is fixed during the navigation process.
The state at a step $t\in[0,T]$ is $\btheta_t, \bt_t, \bz_t$ with a linear state transition:
\begin{equation}
\btheta_{t+1} = \btheta_t+\Delta \btheta_t, ~ \bt_{t+1}=\bt_t + \Delta\bt_t,~ \bz_{t+1}=\bz_{t}+\Delta \bz_t
\end{equation}
We further consider the forward synthesis model as the observation function, and the observation $\bo_t$ at a step $t$ is the synthesized image $\hat{\bI}_t$ combined with the target $\bI$. Given an initial state $\btheta_0, \bt_0, \bz_0$, the agent iteratively takes $T$ steps of actions towards reaching the goal state $\btheta^*, \bt^*, \bz^*$ that can best reproduce $\bI$. At each step, an action $\ba_t\coloneqq \Delta\btheta_t, \Delta \bt_t, \Delta \bz_t$ is taken following the policy $\pi(\ba_t|\bo_t)$.
This formulation leads to a key question: what is a good navigation policy $\pi(\ba_t|\bo_t)$? We now briefly discuss three different strategies in this unified pipeline.

\subsection{Gradient Descent}
\label{sec:gd}
When $G$ is differentiable as considered in this paper, it is straightforward to update the input variables using gradient descent (GD) \cite{chen2020category, yen2020inerf, park2020latentfusion}, yielding the following policy:
\begin{equation}\label{eq: gd_update}
    \pi(\ba_t|\bo_t) = -\lambda \frac{\partial \cL_{p}}{\partial \btheta_t}, -\lambda \frac{\partial \cL_{p}}{\partial \bt_t}, -\lambda \frac{\partial \cL_{p}}{\partial \bz_t} 
\end{equation}
where $\lambda$ controls the speed of gradient descent. $\cL_{p}$ is the perceptual loss~\cite{Johnson2016ECCV} following Chen \etal~\cite{chen2020category}, which measures the discrepancy between the observed image $\bI$ and the synthesized image $\hat{\bI}$. 
Here, the policy $\pi(\ba_t|\bo_t)$ does not contain any trainable parameters, thus can be directly applied without training. On the other hand, the agent may easily get stuck in a local minimum due to the non-convex loss landscape as shown in \figref{fig:teaser}. 
Thus, the final performance highly depends on the initial state $\btheta_0, \bt_0, \bz_0$. The main reason is that the agent cannot look into the future to plan for a long-term reward.

\subsection{Reinforcement Learning}
\label{sec:rl}
In contrast to GD, RL allows the agent to explore the environment to maximize an accumulated reward over multiple steps. The RL policy is inspired by~\cite{Shao2020CVPR}, where RL is adopted for instance-level pose estimation. While this requires a known CAD model, we apply RL for category-level object pose estimation and recover both the object and its pose simultaneously.

Specifically, we apply Soft Actor-Critic to train the RL agent \cite{Haarnoja2018ICML}. As illustrated in \figref{fig:pipeline}, it consists of a policy network $\pi_\psi(\ba_t|\bo_t)$ to produce a stochastic policy and a Q-value function $Q_\phi(\bo_t, \ba_t)$ to inform how good the policy is, with $\psi$ and $\phi$ denoting the parameters of the networks, respectively. 
The policy network is trained to maximize the expected sum of future discounted rewards $\mathbb{E}[\sum_{t=0}^{T}\gamma ^t r_t]$ approximated by $Q_\phi(\bo_t, \ba_t)$, where $T$ is the number of steps, $\gamma$ is a discount factor and $r_t$ is a reward at each step:
\begin{equation}\label{eq:reward}
\begin{aligned}
    r_t = & - \lambda_1\|\bq(\btheta^*-\btheta_t) - \bq(\Delta\btheta_t)\|_2^2 - \lambda_2\| (\bt^*-\bt_t) - \Delta\bt_t \|_2^2 \\
      &   - \lambda_3\|(\bz^*-\bz_t) - \Delta\bz_t \|_2^2
\end{aligned}
\end{equation}
where $\bq(\btheta)$ denotes the quaternion of Euler angles $\btheta$, and the weight parameters $\lambda_1 = 10.0, \lambda_2=5.0, \lambda_3=1.0$ are set to balance the contributions of each term. Here, we assume the target state $\btheta^*,\bt^*,\bz^*$ is available when training the policy network. For example, it can be obtained by randomly sampling a target image from the simulator as $\bI=G(\btheta^*,\bt^*,\bz^*)$. 

There are two major differences comparing RL to GD. Firstly, the reward in \eqref{eq:reward} provides cues for global convergence through direct comparison to the best possible action, while the perception loss in \eqref{eq: gd_update} does not necessarily lead to an update towards reaching the global optimum. Secondly, the RL policy aims to maximize the predicted accumulated reward in future steps, while the GD policy greedily minimizes a one-step loss. Therefore, the RL policy is expected to be less prone to local minima.

\boldparagraph{Training Efficiency} The training of the Soft Actor-Critic is based on an experience replay buffer that stores a set of state-action reward pairs. During early training, unconverged policies can lead an agent to random locations, filling the experience replay buffer with low-reward samples and causing inefficient learning. We improve the training efficiency of RL in two ways.  Following the relabeling strategy  \cite{andrychowicz2017hindsight}, we first relabel the final synthesized image of a failure trial as the target image, turning it into a successful trial. We further manually sample successful trials. 
Adding both, the relabeled and the manually sampled trajectories, to the experience replay buffer allows for speeding up the training. We refer to the supplementary for ablation study of 
the training efficiency.

\subsection{Imitation Learning} 
\label{sec:il}

As expert demonstrations are easily accessible from the simulator, an alternative is to directly learn a policy network via imitation learning. 

\boldparagraph{Behavior Cloning}
Taking the same network structure in RL, we directly train a policy network $\pi_\psi(\ba_t|\bo_t)$ via Behavior Cloning (BC)~\cite{Bojarski2016ARXIV}, see \figref{fig:pipeline}. In this supervised setting, the policy network is trained with one-step supervision, forcing it to reach the goal in one update. The loss can be formulated as follow:
\begin{equation}
\begin{aligned}
    \cL_{IL} = &  \lambda_1\|\bq(\btheta^*-\btheta_t) - \bq(\Delta\btheta_t) \|_2^2 +\lambda_2\| (\bt^*-\bt_t) - \Delta\bt_t \|_2^2 \\
      &  +\lambda_3\|(\bz^*-\bz) - \Delta\bz \|_2^2
\end{aligned}
\label{eq:il}
\end{equation}
where the weight parameters $\lambda_1 = 10.0, \lambda_2=5.0, \lambda_3=1.0$ are the same as \eqref{eq:reward}. 
Again, we assume the target state $\btheta^*,\bt^*,\bz^*$ is available during training.
Although the agent is supervised by the one-step update, it can be applied iteratively to reach the goal during inference.

\boldparagraph{DAgger} 
One disadvantage of BC compared to RL is that the i.i.d. assumption of BC is violated when applied iteratively during inference, as the synthesized image at step $t+1$ depends on previous predictions at step $t$. Therefore, BC suffers from drift when supervised by off-policy data only. This can be avoided by Dataset Aggregation (DAgger)~\cite{Ross2011AISTATS}, which extends the training dataset on-the-fly by collecting data under the current policy. Specifically, given a predicted action $\Delta\btheta, \Delta\bt, \Delta \bz$, the synthesized image $G(\btheta+\Delta\btheta, \bt+\Delta\bt,\bz+\Delta\bz)$ and its corresponding action label are aggregated to the training set. While differing in the training set, the same loss as BC~\eqref{eq:il} is adopted.
Note that DAgger usually requires an expert to annotate such on-policy data. However, in our case, the pose-aware generative model can provide labeled on-policy data for free.

\section{Implementation Details}
\subsection{Pose-Aware Generative Model}
In principle, our navigation framework is compatible with any pose-aware generative model, and we provide two examples in this paper. For both, we use trained networks by the authors and fix them throughout our work.

\boldparagraph{Voxel-based Generative Model} 
Chen \etal \cite{chen2020category} propose a pose-aware generative model using a voxel-based structure. This model is trained in a supervised manner, taking images of single objects and corresponding poses as supervision. Once trained, it could generate images by controlling the camera/object pose and a latent code $\bz$. 

\boldparagraph{GRAF}
Generative Radiance Fields (GRAF) \cite{schwarz2020graf} is a generative model for radiance fields \cite{mildenhall2020nerf} for high-resolution 3D-aware image synthesis. In contrast to the voxel-based generative model, it learns from \textit{unposed} 2D images using the adversarial loss \cite{Goodfellow2014NIPS}. GRAF can generate images conditioned on the camera/object pose and two latent codes for object shape and appearance, respectively. We consider these two latent codes jointly as our $\bz$. Note that adopting 
unsupervised generative models such as GRAF means that the pose estimation can be achieved without collecting posed images. Neither the generative model nor the navigation agent requires real-world posed images for training.

\subsection{Navigation Policy}

\boldparagraph{Architecture} Following Chen \etal~\cite{chen2020category}, we define the rotation Euler angles $\btheta$ by azimuth, elevation and in-plane rotation. The translation $\bt$ is represented by horizontal and vertical shifts in the image plane, and the scale factor along the z-axis which aligns with the principle axis of the camera. Following the official implementations, we set the dimension of $\bz$ to 16 for the voxel-based generator~\cite{chen2020category} and 256 for GRAF~\cite{schwarz2020graf}.  More details about the network architecture of our RL and IL policies can be found in the supplementary. 

\boldparagraph{Inference} For GD, we set the number of update steps $T=50$ and leverage the Adam optimizer~\cite{Kingma2015ICLR} using the same learning rate with Chen \etal\cite{chen2020category}. For RL and IL, we observe that they converge faster and thus use update steps $T=10$. We initialize $\btheta_0,\bt_0$ as the mean pose of all objects within one category. For the latent code, we set $\bz_0=\mathbf{0}$ for all strategies when using the voxel-based generative model, meaning that the agent starts from the mean appearance. We experimentally observe that $\bz$ is harder to estimate for GRAF due to its higher dimension. Thereby we initialize $\bz$ using an encoder similar to Chen \etal~\cite{chen2020category}. Note that GRAF relies on volumetric rendering and is memory intensive when calculating gradients on high-resolution images. Inspired by the patch-discriminator used in GRAF, we use image patches as inputs to the GD policy. RL and IL policies are applied to the full image as both do not back-propagate through the GRAF generator and thus are more memory efficient.

\section{Experiments}

In this section, we first analyze how design choices affect the performance of the navigation strategies in \secref{sec:ablation}. Next, we systematically compare all three strategies in terms of convergence,  robustness and efficiency in \secref{sec:analysis}. Finally, we compare these strategies to state-of-the-art approaches for category-level pose estimation in \secref{sec:sota} on both synthetic and real-world datasets.

\boldparagraph{Dataset}
We first evaluate on \textit{REAL275}~\cite{wang2019normalized}, a standard dataset for benchmarking category-level object pose estimation. We follow the official split of \cite{wang2019normalized} to evaluate on 2760 real-world images, including 6 categories (camera, can, bottle, bowl, laptop and mug). Here, we use the voxel-based generative model provided by Chen \etal~\cite{chen2020category} as our simulator. Note that this voxel-based generator is trained on synthetic images only. To investigate the performance gap between synthetic and real, we also test on synthetic images in one category (laptop) used for training the generator. 

Additionally, we evaluate on \textit{Cars} \cite{dosovitskiy2017CARLA} and \textit{Faces} \cite{liu2015semantic} used in GRAF when using GRAF as the pose-aware generative model. As the poses of the Cars dataset are available, we evaluate on 2000 images used for training GRAF. For the real-world Faces dataset where poses are not available, we sample 2000 images from GRAF as target images for quantitative evaluation and show qualitative evaluation using real-world face images.

We train an RL/IL policy network on each category individually. Since RL and IL are well-known to be sensitive to different random seeds\cite{henderson2018deep}, we apply 10 random seeds and report the standard variation for experiments in Section \ref{sec:ablation} and Section \ref{sec:analysis}. For REAL275, we train on synthetic images used for training the voxel-based generator as their poses are available. As for Cars and Faces, we randomly generate samples from GRAF for training.

\boldparagraph{Metrics}
We follow the evaluation protocol of NOCS~\cite{wang2019normalized} to evaluate average precision ($AP$) at different error thresholds. For the REAL275 dataset where original images contain multiple objects with background, NOCS considers object detection, classification and pose estimation jointly. Following Chen \etal \cite{chen2020category}, we use the trained Mask-RCNN network provided by NOCS to detect and segment objects, such that a single-object target image without background is provided to our visual navigation pipeline. This ensures fair comparison to Chen \etal \cite{chen2020category} and NOCS as all methods rely on the same network for pre-processing.
Following Chen \etal \cite{chen2020category}, the rotation and translation errors are evaluated as:
\begin{equation}
              e_\bR  = arccos \frac{Tr (\bR^* \cdot \bR_T^{-1} ) -1}{2},~~~~
    e_\bt  = \Vert \bt^*- \bt_T \Vert_2
\end{equation}
where $Tr$ represents the trace of a matrix, $\bR_T$ and $\bt_T$ denote the final prediction. 
Following NOCS~\cite{wang2019normalized}, the rotation along the axis of symmetry is not penalized for symmetric object categories, \ie, bottle, bowl and can.

\subsection{How are Policies Affected by Design Choices?}
\label{sec:ablation}

When adopting a trainable policy such as RL or IL, there are several open questions to design choices: Is it necessary to reach the goal in multi-steps? Is it important to recover the latent code $\bz^*$ while we are interested in pose estimation only? We first investigate these questions on the laptop category in REAL275 and its corresponding synthetic images for training the voxel-based generator.

\boldparagraph{Multi-Step v.s. Single-Step}
We investigate whether it is beneficial to perform sequential updates during inference when adopting a trainable policy. Specifically, we compare two variants of IL, behavior cloning (IL$_\text{BC}$) and DAgger (IL$_\text{DA}$). Both variants are supervised by one-step demonstrations during training, while applied for one step or multiple steps during inference. In this comparison, we refer to multi-step as using $T=10$ steps. As for RL that is usually applied for making sequential decisions, we observe a degenerated performance when using single-step and report results in the supplementary.

Fig. \ref{fig:ablation_step} shows rotation $AP$ on both synthetic and real-world target images.  Interestingly, IL$_\text{BC}$ and IL$_\text{DA}$ perform similar at single-step inference. However, their behaviors diverge when taking multiple steps: IL$_\text{BC}$ is degraded while IL$_\text{DA}$ is improved. As IL$_\text{BC}$ is trained with off-policy data, it diverges when applied iteratively. In contrast, IL$_\text{DA}$ overcomes this problem by adding on-policy data. Furthermore, the gap between single-step and multiple-step becomes more prominent when transferred to the real world. It suggests that the multi-step inference helps to overcome the synthetic-to-real gap when leveraging proper training data.

\boldparagraph{Prediction of Latent Code} 
Taking the simple single-step IL$_\text{BC}$ as an example, we  study whether it makes a difference to  recover $\bz^*$ or not. Specifically, we train another IL$_\text{BC}$ policy using the same network architecture but omitting $\|(\bz^*-\bz) - \Delta\bz \|_2^2$ in \eqref{eq:il}.
\figref{fig:il_ablation_1} compares the rotation $AP$ of IL$_\text{BC}$ trained with and without loss on $\Delta\bz$. Surprisingly, adding the loss on $\Delta\bz$ improves the pose estimation accuracy, especially when the target is real-world images. This finding is interesting as iterative update is not applied here, \ie, $\Delta\bz$ does not directly affect the final prediction of $\bR,\bt$.
We hypothesize that recovering $\bz^*$ acts as an auxiliary task which can boost the performance of related tasks~\cite{Zamir2018CVPR}, \ie, the prediction of $\Delta\bR$ and $\Delta\bt$.

\boldparagraph{Discussions} 
We observe that the long-horizon navigation policy can be beneficial despite the trainable policy making reasonable predictions in a single step. However, it is important to take the multi-step inference into account during training, \eg, via on-policy training data in IL. Further, recovering $\bz^*$ acts as a multi-task constraint that helps to improve the performance of pose estimation.

\begin{figure}[t]%
    \centering
    \subfloat[\centering Multi-Step v.s. Single-Step. \label{fig:ablation_step}]{{\includegraphics[width=0.48\linewidth]{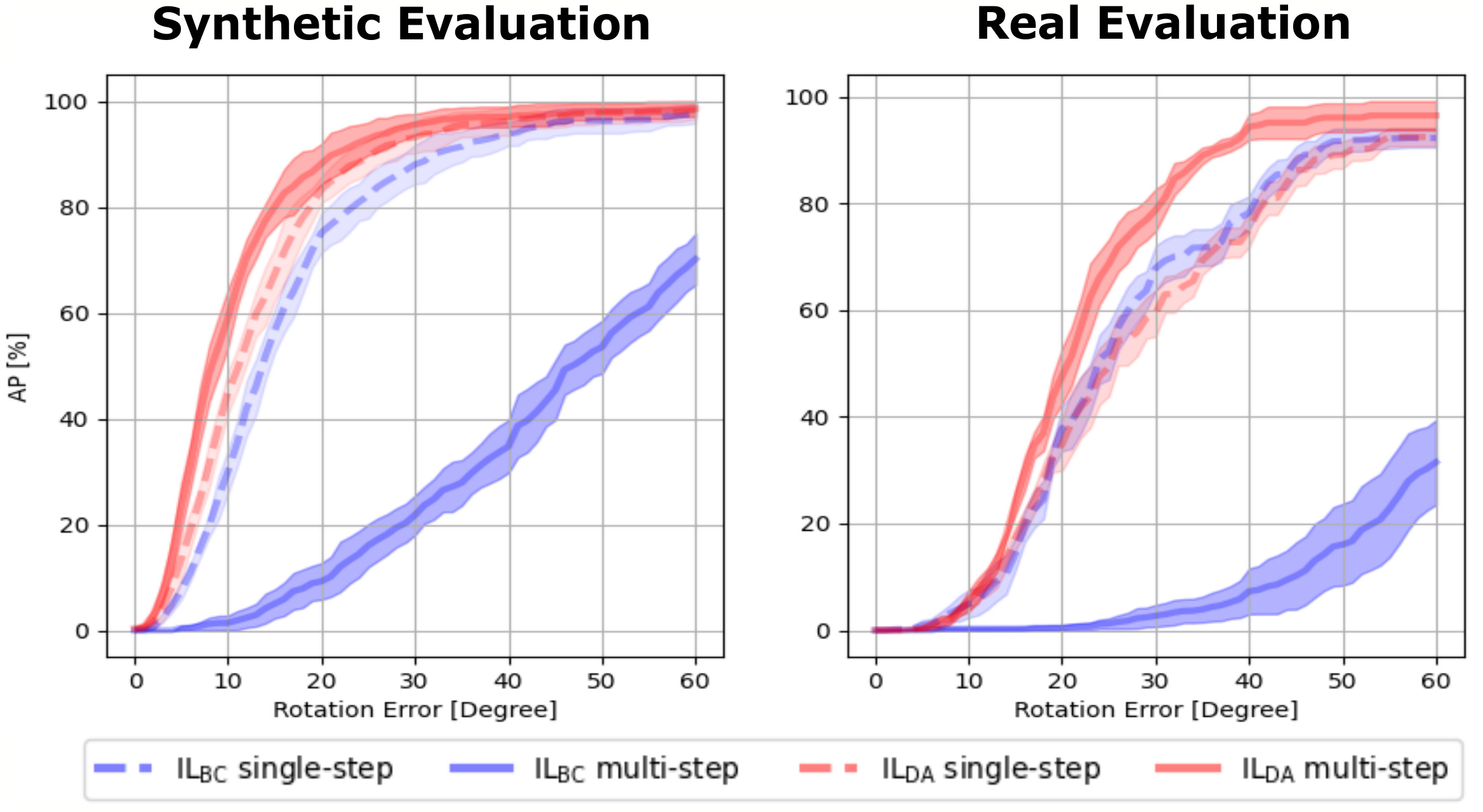} }}%
    \subfloat[\centering Prediction of Latent Code. \label{fig:il_ablation_1}]{{\includegraphics[width=0.48\linewidth]{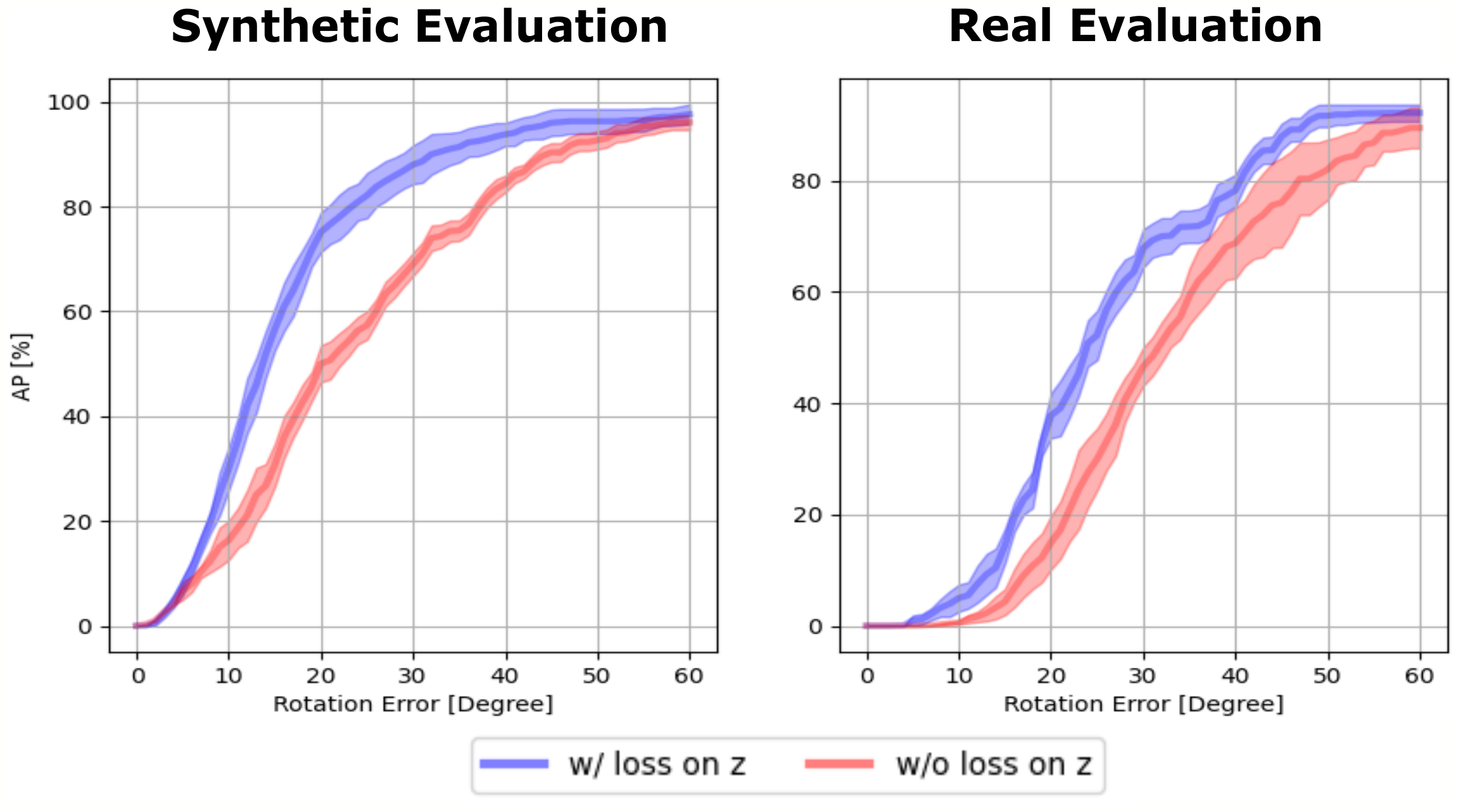} }}%
    \caption{\textbf{Effect of Design Choices.} (a) We compare IL$_\text{BC}$ and IL$_\text{DA}$ on the laptop category on synthetic and real-world images. (b) We compare single-step IL$_\text{BC}$ on the laptop category with and without loss on $\Delta \bz$ on both synthetic and real-world target images.}%
    \label{fig:example}%
\end{figure}

\subsection{What is a Good Navigation Policy?}
We now compare all strategies, GD, RL and IL$_\text{BC}$ and IL$_\text{DA}$. Based on previous analysis, we adopt single-step inference for IL$_\text{BC}$ and multi-step for IL$_\text{DA}$.

\label{sec:analysis}

\begin{figure*}[t]
    \centering
    \begin{tabular}{cc}
    \adjustbox{valign=b}{\begin{tabular}{@{}c@{}}
    \subfloat[Convergence \wrt. initialization.\label{fig:analysis_1}]{%
          \includegraphics[width=.45\linewidth,height=3.8cm]{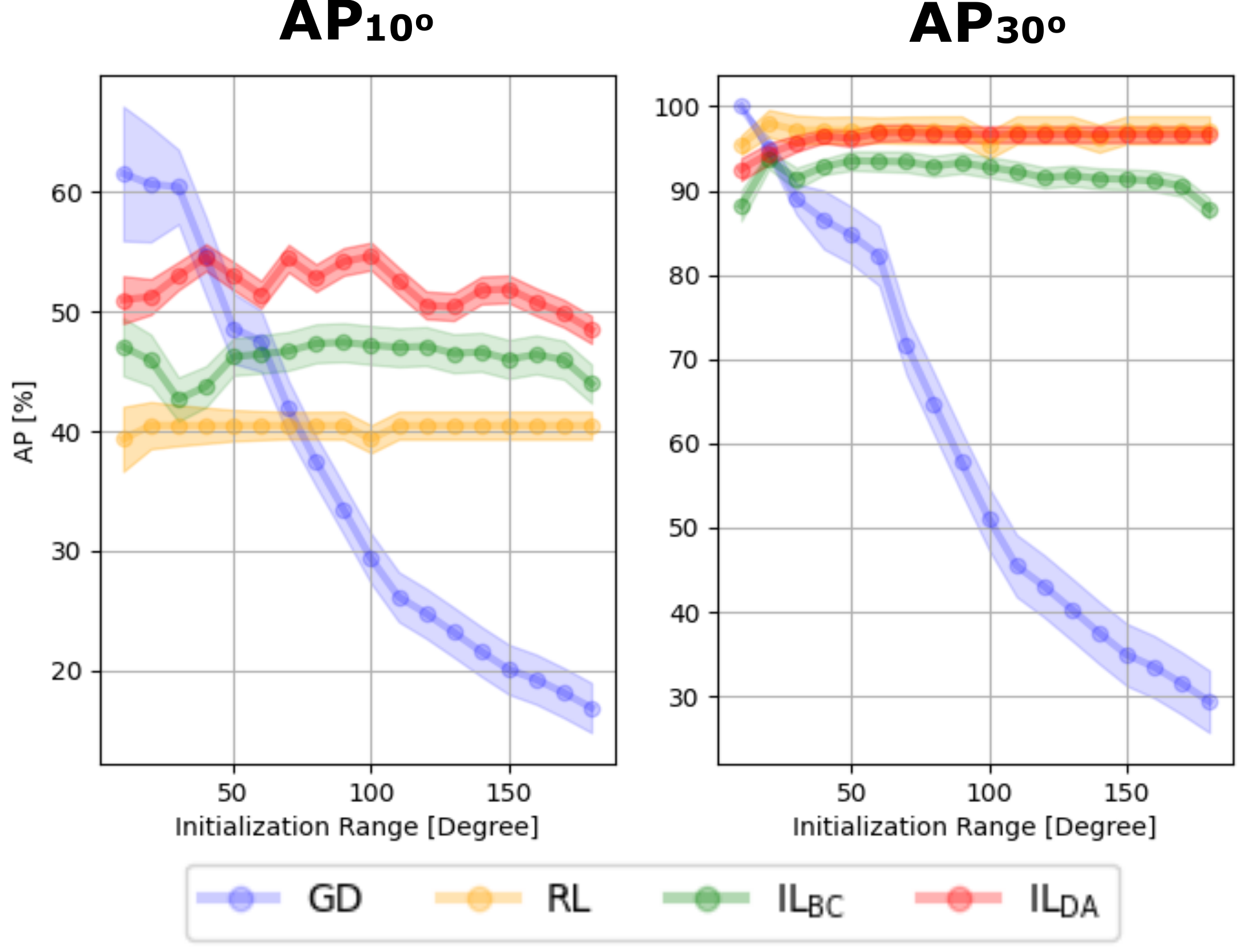}} \\[0.8cm]
          \subfloat[Efficiency.\label{table: time}]{%
        \adjustbox{width=0.45\textwidth}{\begin{tabular}{l|llllll}
\toprule
Strategy       & GD   & GD$_{16}$    &    GD$_{32}$  & RL & IL$_\text{BC}$  &  IL$_\text{DA}$\\ 
\midrule
Training (h) & -- & -- & -- & 25 & 2 & 6       \\ 
Inference (s) & 0.16      & 2.31  & 4.68     & 0.007       & 0.0008 & 0.007      \\
\bottomrule
            \end{tabular}}}
    \end{tabular}} &
    \adjustbox{valign=b}{\subfloat[Robustness.\label{fig: robust}]{%
          \includegraphics[width=.55\linewidth,height=6.1cm]{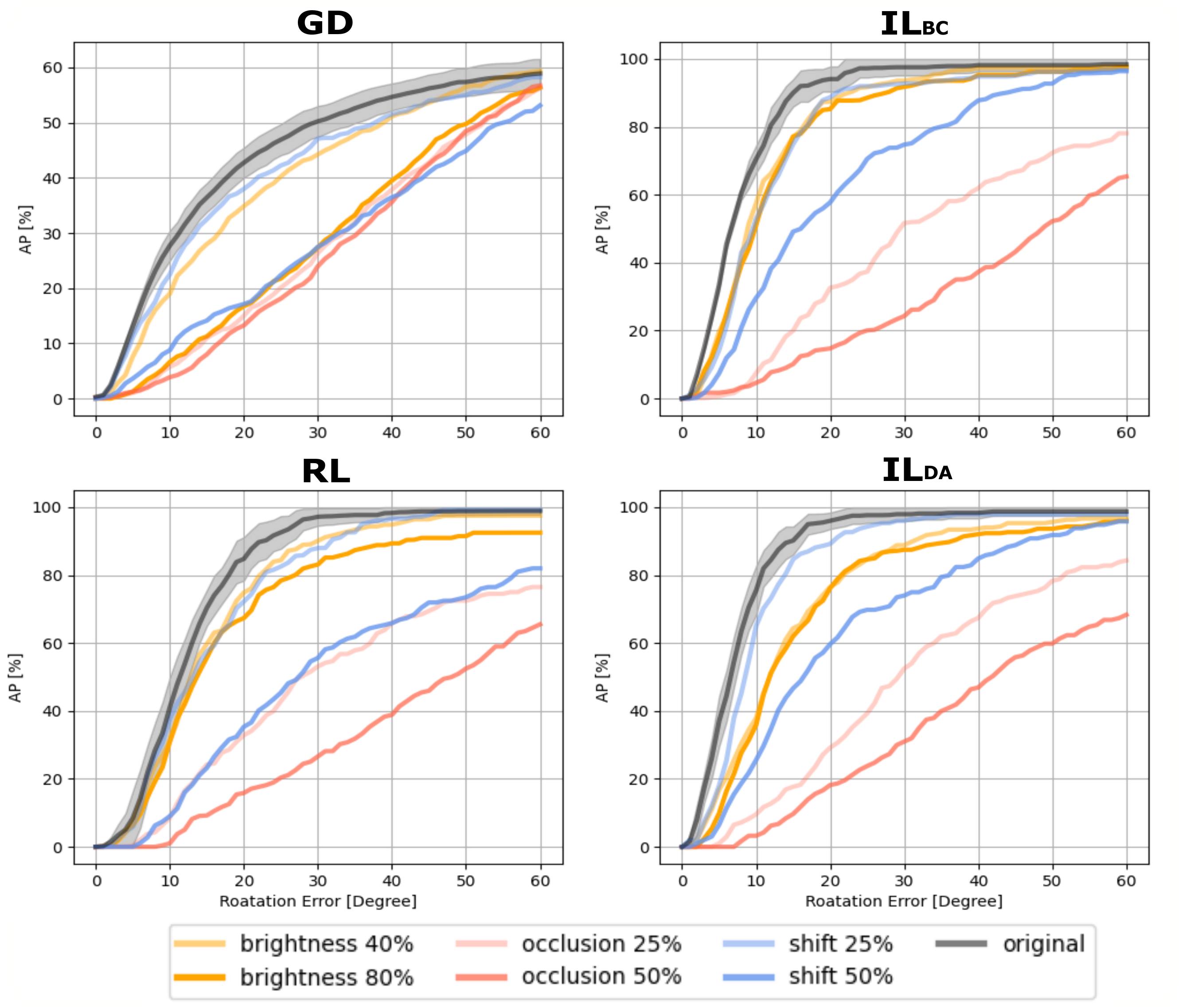}}}
    \end{tabular}
    \caption{\textbf{Convergence, Robustness and Efficiency} of different navigation policies. (a) Rotation $AP_{10\degree}$ and $AP_{30\degree}$ given different initial states. (b) Total training time and averaged inference time on a single image, both evaluated using the voxel-based generative model at the image resolution of $64\times 64$. (c) Rotation $AP$ of  navigation policies against disturbance in brightness, occlusion and shift.}\label{fig & table}
  \end{figure*}

\boldparagraph{Convergence}
We first evaluate GD, RL, IL$_\text{BC}$ and IL$_\text{DA}$ in how the initialization affects the pose estimation. To this goal, we manually control the relative pose between the initial state and the target. We evaluate on Cars where the variation of the target poses is the largest. Specifically, we keep the relative translation fixed, and increase the relative azimuth angle from $10\degree$ to $180 \degree$ with an interval of $10\degree$. We compare the rotation precision $AP_{10\degree}$ and $AP_{30\degree}$ in \figref{fig:analysis_1}. As can be seen,  GD achieves the best precision in the range $[10\degree, 40\degree]$, demonstrating that GD is more precise given a good initialization. This is because that GD is guaranteed to converge to the global optimum of $L_{p}$ given a good initialization. However, the precision of GD degrades significantly when the initialized angle deviates further from the target. In contrast, RL, IL$_\text{BC}$ and IL$_\text{DA}$ are not affected, maintaining almost the same performance in different initialization conditions. Further note that IL$_\text{DA}$ trained with on-policy data achieves superior performance compared to RL while IL$_\text{BC}$ is less competitive (at $AP_{30\degree}$).

\boldparagraph{Robustness} 
Inspired by Chen \etal \cite{chen2020category}, we compare the robustness of GD, RL, IL$_\text{BC}$ and IL$_\text{DA}$ against variations of brightness, occlusion and shift in the target image. This is evaluated on the synthetic images of the laptop category. As shown in Fig. \ref{fig: robust}, GD is more robust against disturbance in brightness and shift compared to other strategies. One possible explanation is that the perceptual loss is more robust, as it is calculated based on VGG~\cite{simonyan2014very} pretrained on ImageNet~\cite{Deng2009CVPR}. In contrast, both RL and IL policy networks are trained on synthetic images only, thus being less robust against the domain shift. All methods struggle to some extent in terms of occlusions,  which can be a disadvantage of the analysis-by-synthesis pipeline. Note that neither RL nor IL policy is trained with data augmentation regarding brightness, occlusion, or shift. We expect better robustness when trained with augmentation against these disturbances.

\boldparagraph{Efficiency}
Fig. \ref{table: time} shows the training and inference time of different strategies on the same device NVIDIA RTX 3090. For GD, we further evaluate the inference time using multiple different initial states. This strategy is used in Chen \etal \cite{chen2020category} to avoid converging to local minima. Despite that GD does not require training, its inference takes longer compared to RL, IL$_\text{BC}$ and IL$_\text{DA}$. Furthermore, the time cost of GD increases \wrt the number of initial states. Taking 32 initial states as used in Chen \etal \cite{chen2020category} requires almost 7 seconds for one target image. For trainable policies, the training time of both IL variants is less compared to the RL policy thanks to the direct supervision. During inference, IL$_\text{BC}$ is the most efficient method as we take only a single-step update when using IL$_\text{BC}$. RL and IL$_\text{DA}$ take longer, but the overhead is acceptable.

\begin{figure*}[t]
\centering
\includegraphics[width=0.9\linewidth]{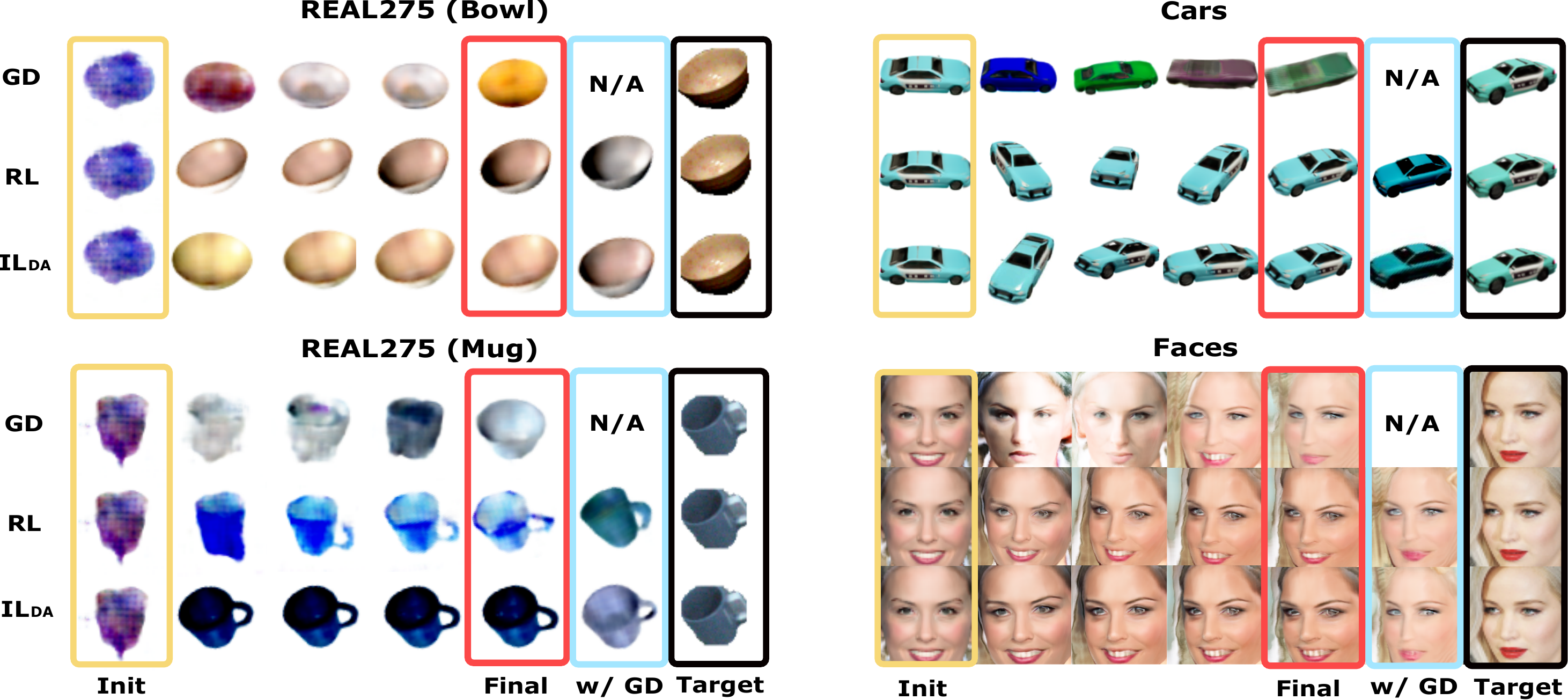}
\caption{\textbf{Qualitative Comparison} of different strategies, including the initialization, the optimization process, the final image and the target image. For RL and IL$_\text{DA}$, we further show the synthesized  image after adding 10 steps of GD. }
\label{fig: visual_result}
\vspace{-10pt}
\end{figure*}

\boldparagraph{Discussions}
Our analysis shows that GD achieves the best precision given a good initialization and better robustness against brightness and shift. However, it easily gets stuck in local minima when the initial state is far from the target. Solving this problem by adopting multiple initial states sacrifices efficiency. On the other hand, RL and IL policies are efficient and less prone to local minima. When augmented with on-policy data, IL$_\text{DA}$ performs similar or even better compared to RL while requiring less training time. Therefore, we suggest to combine GD's precision and robustness with the convergence and efficiency of IL$_\text{DA}$ by applying a few steps of GD (\eg, $T=10$) after IL$_\text{DA}$. We show results of this simple hybrid method in the next section.

\subsection{Comparison to the State-of-the-Art}
\label{sec:sota}

 \boldparagraph{Baselines} We now compare different strategies to baselines for category-level object pose estimation. We first evaluate a simple baseline following Chen \etal \cite{chen2020category}, where a VGG16~\cite{simonyan2014very} is adopted to regress the object pose from an RGB image directly. We then compare to state-of-the-art analysis-by-synthesis approaches, including iNeRF\cite{yen2020inerf} and Chen \etal \cite{chen2020category}. Both approaches follow the principle of the GD policy but improve from different aspects:
 iNeRF samples image patches in interested regions to calculate the loss while \cite{chen2020category} starts from 32 different initial states. We also consider NOCS \cite{wang2019normalized} as a reference on the REAL275 dataset. Note that NOCS does not apply to Cars and Faces due to the lack of supervision. Moreover, NOCS is trained jointly on synthetic and real-world RGB-D images, while our method is trained on synthetic RGB images only.

\begin{table}[t]
\scriptsize %
\renewcommand\theadfont{\setfsize}
\resizebox{\linewidth}{!}{%
\renewcommand{\arraystretch}{1.1}
\begin{tabular}{@{}C{2cm}C{1cm}|C{1cm}|C{1cm}C{1cm}C{1cm}C{1cm}C{1cm}C{1cm}C{1cm}C{1cm}@{}}
\toprule

Dataset & Metric  &\makecell{NOCS* \\  \cite{wang2019normalized}}&
  \makecell{VGG \\ \cite{simonyan2014very}} & \makecell{iNeRF \\ \cite{yen2020inerf}} & \makecell{Chen\\ \cite{chen2020category}} & \textbf{GD} & \textbf{\makecell{RL }} & \textbf{\makecell{IL$_\text{DA}$ }} & \textbf{\makecell{RL  \\ w/ GD}} & \textbf{\makecell{IL$_\text{DA}$ \\ w/ GD}}   \\ \cmidrule(l){1-11}
 \multicolumn{1}{c}{\multirow{6}{*}{\textbf{\makecell{REAL275  \\ Dataset \\ (Symmetry)}}}}   & $AP_{10 \degree}$   & 32.8 &   6.4  & 21.0   & 24.0&20.5& 18.6&21.6&\underline{24.8} & \textbf{25.0}\\
\multicolumn{1}{c}{} & $AP_{30 \degree}$    & 66.5  &  34.8 & 88.7   &  92.1 &   86.2 & 91.7 & \underline{93.6}&  92.5& \textbf{94.2}\\
\multicolumn{1}{c}{} & $AP_{60 \degree}$  &99.3  &  76.3 &97.1  &\textbf{99.9}&   96.7 &98.8&\underline{99.6}& \underline{99.6} & \textbf{99.9}\\  
\multicolumn{1}{c}{} & $AP_{5cm}$   & 93.4  &  7.8 & 11.9  & 12.7 &   11.9&11.8 &12.4& \underline{13.2} & \textbf{14.6}\\ 
\multicolumn{1}{c}{} & $AP_{10cm}$    &95.0 & 23.7 & 26.1   &  27.4 &   24.5&23.9&\textbf{29.1}& 27.2& \underline{28.8}\\ 
\multicolumn{1}{c}{} & $AP_{15cm}$   & 97.3 &  38.1  & 43.8   & \textbf{46.9} &   41.4& 39.5 & 42.3& 42.6& \underline{46.4}\\ \hline 
\multicolumn{1}{c}{\multirow{6}{*}{\textbf{\makecell{REAL275  \\ Dataset \\ (Asymmetry)}}}} & $AP_{10 \degree}$  &  20.5 & 0.6   & 5.1 &\textbf{6.9} &5.0 & 5.1  & 4.8& 6.5& \underline{6.8}\\
\multicolumn{1}{c}{} & $AP_{30 \degree}$     & 55.5 & 12.4 & 43.1 &\underline{59.5} &   21.1 & 51.6& 53.5 & 58.7& \textbf{60.0}\\
\multicolumn{1}{c}{} & $AP_{60 \degree}$ &93.3 &  35.1 & 62.8  &79.2   &  35.0   & 74.5& \underline{80.6}& 76.3 &\textbf{82.3}\\  
\multicolumn{1}{c}{} & $AP_{5cm}$      &87.7& 10.3& 7.7 & 12.1 &   9.8& 9.7& \textbf{17.5}& \underline{12.8}& 12.5 \\ 
\multicolumn{1}{c}{} & $AP_{10cm}$     &98.2 & 38.1& 33.2  &   42.4 &   27.7 & 41.7& \textbf{52.2}& 42.2& \underline{46.8}\\ 
\multicolumn{1}{c}{} & $AP_{15cm}$    & 99.5 & 61.8& 48.7  & \underline{73.1} & 50.6& 71.6& \textbf{75.8}& 73.0&72.8 \\  \hline 
\multicolumn{1}{c}{\multirow{6}{*}{\textbf{\makecell{Cars \\ Dataset}}}}   & $AP_{10 \degree}$  &$\setminus$  &   5.6  & 31.7   & 51.8&21.3&42.4 & 47.6 & \underline{62.9} &\textbf{65.3}\\
\multicolumn{1}{c}{} & $AP_{30 \degree}$    &$\setminus$ &  15.4  & 45.4   & 85.5& 33.7&92.8 & 93.5&\underline{93.8} &\textbf{94.1}\\
\multicolumn{1}{c}{} & $AP_{60 \degree}$   &$\setminus$  &  32.8 & 56.6   & 93.7&   37.4  & 94.2&\underline{98.2} &97.7 & \textbf{98.8}\\  
\multicolumn{1}{c}{} & $AP_{1cm}$    &$\setminus$ &  8.9 & 12.4   & \underline{35.7}& 9.6&27.2  & 35.5& 29.1& \textbf{36.7}\\ 
\multicolumn{1}{c}{} & $AP_{3cm}$    &$\setminus$  &35.2& 41.6   & \underline{75.8}& 32.7&71.8 &\textbf{76.3}& 72.5& 75.6\\ 
\multicolumn{1}{c}{} & $AP_{6cm}$   &$\setminus$ &  52.1 & 68.3   & 85.7&   60.1 &91.8 & \underline{92.0}& 91.4& \textbf{92.9}\\ \hline 
\multicolumn{1}{c}{\multirow{6}{*}{\textbf{\makecell{Faces \\ Dataset}}}} & $AP_{5 \degree}$  &$\setminus$ &   5.3   & 4.6   & 24.8&2.0&17.3 &\textbf{25.8} & 15.4& \underline{23.1}\\
\multicolumn{1}{c}{} & $AP_{15 \degree}$  &$\setminus$  &  32.8   & 42.6   &88.7&   35.9&84.2& \underline{89.5}& 88.4& \textbf{90.8}\\
\multicolumn{1}{c}{} & $AP_{30 \degree}$   &$\setminus$ &  71.1  & 80.9  &92.5&   81.2&98.6 &98.3 &\textbf{99.5}& \underline{99.1}\\  
\multicolumn{1}{c}{} & $AP_{1cm}$   &$\setminus$  & 11.0& 18.2   & \underline{27.4}&  14.3&\textbf{27.9} & 25.9& 26.7& 25.3\\ 
\multicolumn{1}{c}{} & $AP_{3cm}$   &$\setminus$ & 41.0& 59.6   & \textbf{92.6}&  53.8 &86.3 & 90.1&87.8 &\underline{91.5}\\ 
\multicolumn{1}{c}{} & $AP_{6cm}$   &$\setminus$ & 72.9  & 85.7  & \underline{98.5}&   82.3&97.2 & 97.7& \underline{98.5}& \textbf{99.5}\\ \hline

\multicolumn{1}{c}{\multirow{2}{*}{\textbf{\makecell{Mean}}}} & $AP_\text{rot}$   &$\setminus$&27.4&48.3&66.6&39.7&64.2&67.2&\underline{68.0}&\textbf{70.0}\\
\multicolumn{1}{c}{} & $AP_\text{tran}$   &$\setminus$ &33.4 &38.1&52.5&34.9&50.0&\textbf{53.9}&51.4&\underline{53.6}\\
\bottomrule
\end{tabular}

}
\vspace{1pt}
\scriptsize{*NOCS is based on RGB-D while the others are based on RGB images.}
\vspace{1pt}
\textcolor{black}{\caption{\textbf{Quantitative Comparison} of category-level pose estimation on  different datasets.}
\label{table: all_results}}
\vspace{-17pt}
\end{table}

\boldparagraph{Results}
The quantitative results are shown in \tabref{table: all_results}.
Firstly, we observe that VGG is outperformed by other methods, suggesting that it is difficult to regress the pose from a single RGB image directly. For iNeRF, it is interesting that it outperforms GD based on interest region sampling, even when  GD is applied to the full image on REAL275. However, it struggles to overcome the convergence problem given the uncurated initialization range as in this paper, particularly on the Cars dataset where the initial azimuth is within the range of $[-180\degree, 180\degree]$. Chen \etal~\cite{chen2020category}  significantly outperforms GD by leveraging 32 initial states, 
but is time-consuming as shown in \figref{table: time}.  

In contrast, RL and IL$_\text{DA}$ achieve competitive performance compared to \cite{chen2020category} but are remarkably more efficient. 
Moreover, the simple hybrid approaches, RL w/ GD and IL$_\text{DA}$ w/ GD, often lead to better performance. As RL/IL$_\text{DA}$ provides a fairly good start, the subsequent GD converges in only a few steps. This reduces the required steps of ``w/ GD'' to $T=10$ in contrast to $T=50$ in the standard GD, and the inference time of RL, IL$_\text{BC}$, IL$_\text{DA}$ change to 0.033s, 0.027s, 0.033s respectively. The hybrid approach is still much more efficient compared to  
\cite{chen2020category} (0.033s v.s. 4.68s). Note that IL$_\text{DA}$ w/ GD yields the best performance among all RGB based approaches and sometimes even achieves comparable performance to NOCS based on RGB-D input. 
Interestingly, the hybrid approach sometimes worsens the translation performance, \eg, on the asymmetry categories of REAL275. This might be due to the scale ambiguity of the generative model. Here, NOCS achieves superior performance in translation leveraging depth maps. Lastly, it is worth noting that our learned policies consistently improve the inference of different pose-aware generators. This brings hope to apply our method to more advanced synthesis models for more challenging tasks.

We further show the qualitative comparison of different navigation strategies in \figref{fig: visual_result}. Note that RL and IL both allow for converging to the correct pose starting from a bad initialization, \eg, the car example. Furthermore, adding 10 steps of GD helps to refine the object pose, see the mug category of REAL275. More  qualitative comparisons are provided in the supplementary.

\section{Conclusions}
In this paper, we formulate the category-level object pose estimation problem as a long-horizon visual navigation task.
We experimentally analyze three different navigation policies in terms of convergence, robustness and efficiency. Based on our analysis, we come up with a simple yet effective hybrid approach that enhances the convergence of existing analysis-by-synthesis approaches without sacrificing the efficiency. We further show that it improves the inference of different pose-aware generative models. However, the scale ambiguity of monocular images remains unsolved, thus estimating correct translation is particularly challenging. We plan to tackle these challenges in the future.

\small{\boldparagraph{Acknowledgement} This work is supported in NSFC under grant U21B2004, and partially supported by Shenzhen Portion of Shenzhen-Hong Kong Science and Technology Innovation Cooperation Zone under HZQB-KCZYB-20200089, the HK RGC under T42-409/18-R and 14202918, the Multi-Scale Medical Robotics Centre, InnoHK, and the VC Fund 4930745 of the CUHK T Stone Robotics Institute.}

\bibliographystyle{splncs04}
\bibliography{bibliography_long,egbib}
\end{document}